# Daily Forecasting for Annual Time Series Datasets Using Similarity-Based Machine Learning Methods: A Case Study in the Energy Market

## Mahdi Goldani

ASSOCIATE PROFESSOR OF ECONOMICS, HAKIM SABZEVARI UNIVERSITY, IRAN EMAIL: M.GOLDANI@HSU.AC.IR

## Abstract

**Purpose:** The policy environment of countries changes rapidly, influencing macro-level indicators such as the Energy Security Index. However, this index is only reported annually, limiting its responsiveness to short-term fluctuations. To address this gap, the present study introduces a daily proxy for the Energy Security Index and applies it to forecast energy security at a daily frequency.

**Methods:** The study employs a two-stage approach: first, a suitable daily proxy for the annual Energy Security Index is identified by applying six time series similarity measures (DTW, Soft-DTW, LCSS, EDR, and Hausdorff distance) to key energy-related variables. Second, the selected proxy is modeled using the XGBoost algorithm to generate 15-day-ahead forecasts, enabling high-frequency monitoring of energy security dynamics.

**Main findings:** As the result of proxy choosing, *Volume_Brent* consistently emerged as the most suitable proxy across the majority of methods. The model demonstrated strong performance, with an R² of 0.981 on the training set and 0.945 on the test set, and acceptable error metrics (RMSE = 1354.89, MAE = 800.14). The 15-day forecast of Brent volume indicates short-term fluctuations, with a peak around day 4, a decline until day 8, a rise near day 10, and a downward trend toward day 15, accompanied by prediction intervals.

**value-added contribution:** By integrating time series similarity measures with machine learning–based forecasting, this study provides a novel framework for converting low-frequency macroeconomic indicators into high-frequency, actionable signals. The approach enables real-time monitoring of the Energy Security Index, offering policymakers and analysts a scalable and practical tool to respond more rapidly to fast-changing policy and market conditions, especially in data-scarce environments.

Keywords: Time Series Similarity, High-Frequency Proxy, Energy Security Index, XGBoost Forecasting

**Introduction**

High-frequency data are more informative, detailed, and timely than their low-frequency counterparts and thus are essential for detailed analysis in economics, finance, and energy research. The fine temporal resolution catches short-term variability, seasonality, and abrupt shocks that are missed by annual or quarterly data, considerably improving forecasting and dynamic modeling (Song et al., 2024; Bolivar, 2025; Zhang et al., 2005).

Most macroeconomic and policy-relevant statistics, nevertheless, are compiled quarterly or annually, which poses a challenge to real-time monitoring, nowcasting, simulation modeling, and policy making (Sihombing et al., 2025). Annual data especially do not capture short-run volatility, thereby decreasing adaptability and responsiveness of forecasting models (Giannone et al., 2008).

To overcome this, various temporal disaggregation techniques have been suggested. Classical econometric methods like Denton, Chow–Lin, and Litterman redistribute low-frequency series to higher-frequency values based on statistical assumptions and high-frequency indicators (Feldkircher et al., 2021). More recent advances feature computational software like the Python package tempdisagg, which combines classical methods with improvements like automatic parameter tuning, ensemble modeling, and missing-data handling (Vera-Jaramillo, 2025). Additionally, novel approaches such as the LASSO-based extension of Chow–Lin enable disaggregation in high-dimensional setups, facilitating the utilization of large sets of indicators for producing sparse and interpretable estimates (Mosley et al., 2022).

In spite of these developments, current methods assume the presence of high-frequency indicators, an assumption that fails in many fields. In applications like energy security analysis, for example, where official measurements are only released at annual or infrequent intervals, researchers do not have timely inputs for forecasting and nowcasting. This results in a fundamental temporal mismatch between data availability and the need for real-time analysis. Although proxy construction is occasionally tried through conventional econometric or machine learning feature selection (Wilkinson & Pickett, 2006; Hastie et al., 2009), systematic methods for constructing validated high-frequency proxies from low-frequency indicators are relatively underdeveloped.

The current study seeks to fill this gap through the application of time-series similarity methods, including Dynamic Time Warping (DTW), Euclidean distance, correlation-based similarity, and autocorrelation profiles, to detect and validate high-frequency proxy variables for annual energy security indicators (Din & Shi, 2017). By emphasizing temporal dynamics over absolute values, the suggested approach enables the generation of daily-level proxy series that successfully track trends, seasonal patterns, and external shocks. In a demonstration case study, we develop a daily energy security proxy that is a responsive and high-resolution measure for monitoring, forecasting, and policy assessment. The approach offers a practical way forward for undertaking real-time analysis, risk monitoring, and strategic planning in volatile energy markets even where high-frequency official statistics are not immediately available.

**Methodology**

**Dataset**

This study aims to predict the energy security index at a daily frequency. In terms of daily prediction for the annual energy security index, a daily proxy for the energy security index is required. To accomplish this objective, the data set consisting of daily frequency data reported in the energy sector is used. Energy prices, including crude oil prices, natural gas prices, electricity prices (spot or wholesale), and coal prices, have been extracted from Yahoo Finance using its Python library. "The data frequency is daily, ranging from January 1st, 2011 to December 30th, 2023.

The WEC publishes a World Energy Trilemma Index, which assesses energy security along with sustainability and equity. It ranks as countries based on their energy policies and outcomes. The index includes components related to energy security, such as energy availability, affordability, and sustainability. The energy security data for Iran from 2011 to 2023 has been extracted from this source. Iran had an overall score of 59.8 and a rank of 61 in the 2023 World Energy Trilemma Index. The country scored 59.8 in Energy Security, 81.1 in Energy Equity, and 47.7 in Environmental Sustainability (figure 1).

Figure1. 2023 Energy Trilemma Index in Iran

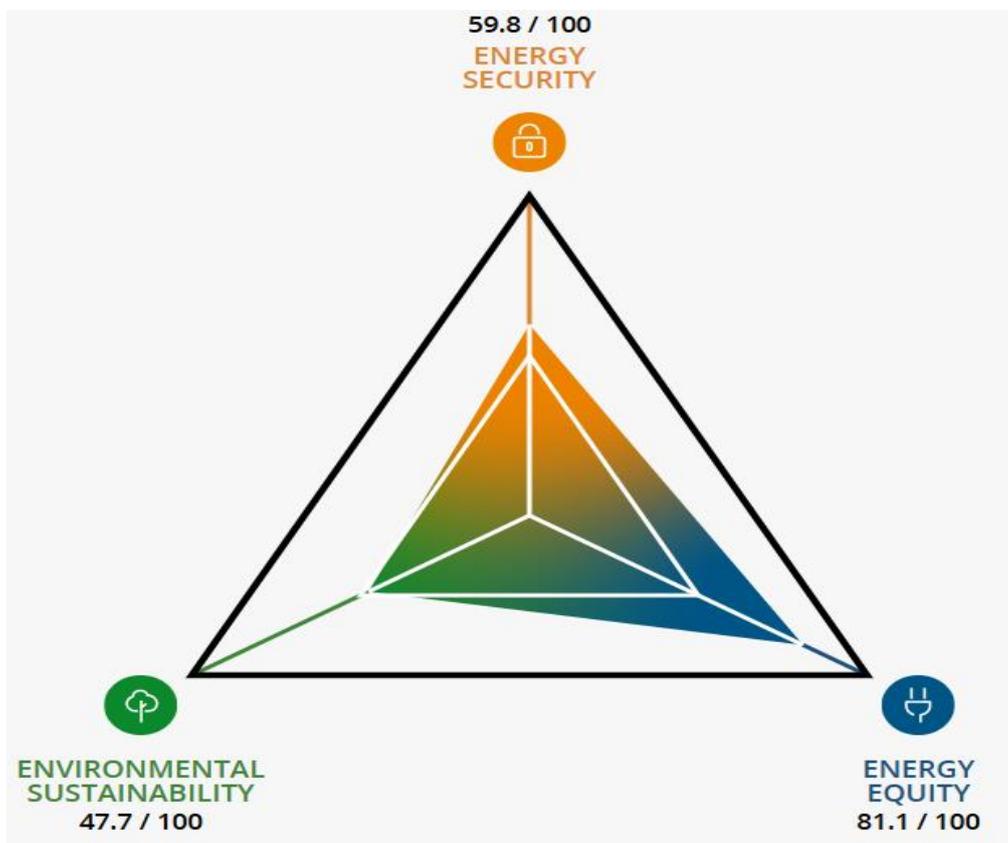

Source: WEC, 2023

**Data preprocessing**

Given the nature of time series data and the behavioral complexities of macro variables, this study employed an advanced approach for data preprocessing. Specifically, to handle missing values, an Autoencoder-based imputation method was used, which is one of the deep learning techniques applied in data reconstruction. This method can more accurately reconstruct missing values by learning the internal structure of the data. To enhance the performance of the neural networks in this process, the data were standardized using the StandardScaler algorithm prior to being fed into the model, thereby minimizing the impact of differences in variable scales on the model results.

**Methodology**

    a. **Choosing proxy**

Time series similarity methods have been employed to identify the best proxy among the daily data for energy security. Before applying the similarity methods, the frequency of the data should be the same. Therefore, the daily data frequency has been converted into an annual frequency by calculating the average for each year.

Time series similarity techniques are used to detect dynamic patterns among countries, particularly if the data possesses temporal dependencies, nonlinear trends, or potential phase or scale shifts. The techniques generally fall into one of three classes: point-to-point measures, elastic measures, and geometric measures. Point-to-point measures, including Euclidean distance and Pearson correlation, directly compare values at each time step and require perfectly aligned series of identical length. Elastic measures, including Dynamic Time Warping (DTW), LCSS, EDR, ERP, and TWED, are more flexible and stretch sequences with different timing or speed, thus being more resistant to phase shifts, irregular sampling, and noise. Geometric measures, including Fréchet, Hausdorff, and SSPD distances, view time series as trajectories in space and compute similarity from the overall shape and structure of their trajectory. Each of these methods has special strengths in encoding piecewise temporal relations and structural patterns within complex time series data. The DTW Distance, Soft-DTW Distance, LCSS, EDR, and Hausdorff distance have been used to measure the similarity in this study.

DTW is a robust measure for similarity between two series. It has the capability to align sequences so points with similar patterns are mapped even though they occur at different times (figure 1). This elastic warping allows DTW to be used for comparison of time series of different lengths. Additionally, the resulting distance is very robust to outliers (li, 2021). DTW is widely used in theme discovery (Din and Shi, 2017), indexing (Rakthanmanon et al, 2012), gesture recognition (Zadghorban and Nahvi, 2018) etc.

Figure2. Dynamic Time Warping Alignment

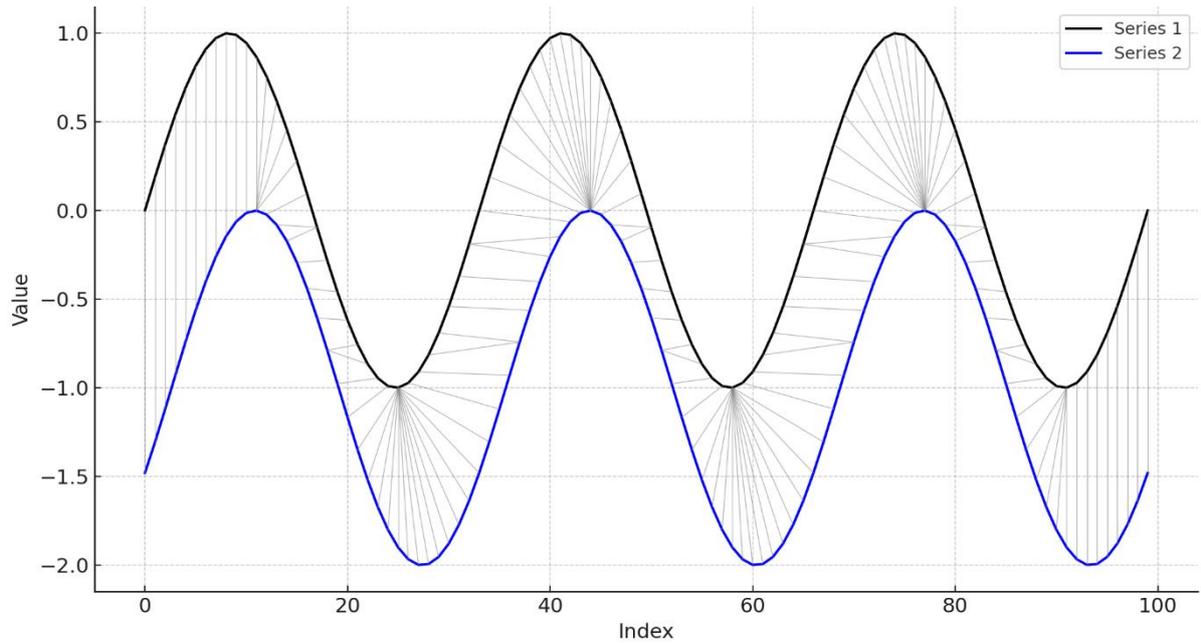

A longest common subsequence (LCS) refers to the longest sequence that appears in the same relative order (but not necessarily contiguously) in all sequences within a given set—typically two sequences. Unlike substrings, subsequences do not require the elements to be adjacent in the original sequences. Determining the LCS is a well-known problem in computer science, forming the foundation of data comparison tools like the *diff* utility. It has significant applications in fields such as computational linguistics and bioinformatics and is also extensively used by version control systems like Git to merge and reconcile changes across different versions of files.

The Longest Common Subsequence (LCS) problem is typically solved using dynamic programming. The standard formula for computing the LCS of two sequences X and YYY, of lengths mmm and n respectively, uses a 2D table L where L[i][j] holds the length of the LCS of the prefixes X [0, i−1] and Y [0, j−1].

Let $X = x_1, x_2, \ldots, x_m$ and $Y = y_1, y_2, \ldots, y_n$.

Define L[i][j] as length of LCSS of X [0, i−1] and Y [0, j−1].

$$L[i][j] = \begin{cases} 0 & if\ i = 0\ or\ j = 0 \\ l[i-1][j-1] + 1 & if\ x_i = y_i \\ \max(L[i-1][j], L[i][j-1]) & if\ x_i \neq y_i \end{cases} \quad (1)$$

Originally introduced to assess the similarity between character strings, edit distances have also been effectively applied to measuring trajectory similarity. The core idea behind edit distance is to quantify how many modifications (or "edits") are needed to transform one trajectory into another for example, by removing a data point. Each of these edits incurs a certain cost. A notable variation, proposed by Chen, Özsu, and Oria (2005), is called Edit Distance on Real sequences (EDR). In EDR, each edit operation, either deleting a point or matching two dissimilar points has a fixed, unit cost (Tao et al, 2021).

The edit distance on real sequence (EDR) of A and B is defined as:

$$EDR(A,B) = \begin{cases} n & \text{if } B \text{ is empty} \\ m & \text{if } A \text{ is empty} \\ \min( \\ EDR(A_{[2,n]}, B_{[2,m]}) + penalty(a_1, b_1) \\ EDR(A, B_{[2,m]}) + 1 \\ EDR(A_{[2,n]}, B) + 1) & \text{otherwise} \end{cases} \quad (2)$$

where $penalty(a_1, b_1)$ is 0 if $dist_\infty(a_1, b_1) < \epsilon$ or 1 otherwise.

The Hausdorff distance, named after Felix Hausdorff, measures how far two sets are from each other by identifying the greatest distance from a point in one set to the closest point in the other set. More formally, consider two sets of spatial points:

A= $[a_1, a_2, ..., a_n]$ and B= $[b_1, b_2, ..., b_m]$ where n and mmm represent the number of points in sets A and B, respectively. The Hausdorff distance h (A, B) is then defined as the maximum of all the shortest distances from a point in one set to any point in the other set (Makris et al, 2021).

Figure3. The Hausdorff distance H (A, B) between A and B.

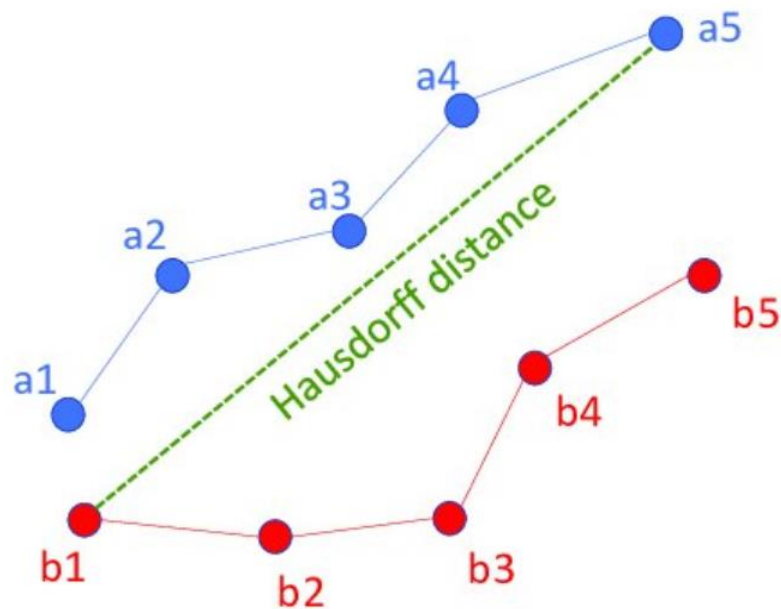

Source: Makris et al(2021)

**XGBoost model**

This study adopts the XGBoost algorithm due to its outstanding capabilities, including high predictive accuracy, computational speed, resistance to overfitting, and adaptability (Sun et al,

2023) Extreme Gradient Boosting (XGBoost) is an advanced gradient boosting framework designed for efficiency, scalability, and portability, making it widely popular in both competitive machine learning and real-world applications. At its core, gradient boosting constructs models incrementally by combining a series of weak learners—typically decision trees—each one trained to reduce the residual errors made by the previous model. XGBoost enhances this framework by incorporating techniques that improve both prediction accuracy and computational efficiency. Specifically, it applies L1 (Lasso) and L2 (Ridge) regularization to avoid overfitting. These regularization methods penalize overly complex models, encouraging simplicity and better generalization to unseen data (Chen, 2016). A standout feature of XGBoost is its ability to perform computations in parallel, significantly accelerating the training process. Unlike traditional boosting methods, XGBoost builds decision trees simultaneously using multiple CPU cores, which increases speed and efficiency (Friedman, 2001). Another important advantage is XGBoost's capability to manage missing data internally. Rather than requiring manual imputation, it learns the best way to handle missing values during training. This simplifies data preprocessing and enhances model reliability, especially when working with imperfect real-world datasets (Chen, 2014).

To avoid over-complex models, XGBoost uses a depth-first pruning strategy, controlling tree growth through parameters such as maximum tree depth and a minimum threshold for further splits. This pruning mechanism effectively mitigates overfitting and boosts the model's generalization ability (Chen, 2016). XGBoost is also optimized for sparse data structures, efficiently handling datasets with missing or zero values by recognizing and exploiting sparsity patterns. This is especially beneficial for large-scale applications where such data characteristics are common. Furthermore, the algorithm offers customization options that allow users to define objective functions and evaluation metrics tailored to specific problems. This flexibility makes XGBoost suitable for a broad range of tasks, including regression, classification, and ranking. To fine-tune the model's performance, grid search was employed to explore various hyperparameter combinations. This method evaluates each parameter set through cross-validation, ultimately selecting the configuration that offers the best balance between predictive accuracy and model simplicity(Ke et al, 2017).

**Result**

We aim to estimate the daily dynamics of the annual Energy Security Index, a critical measure reflecting the stability and resilience of the energy sector. Since the index is reported on an annual basis, this study proposes a new approach to address this challenge. The approach involves identifying a robust proxy for energy security that is reported daily. By examining the behavior of this variable as a proxy for energy security, we are able to predict the future daily trend of energy security and its daily fluctuations.

**Identification of a Daily Proxy for Energy Security**

Six time series similarity methods are applied to identify the variable most similar to the Energy Security Index. These methods include DTW Distance, Soft-DTW Distance, LCSS, EDR, and Hausdorff distance. In these methods, the frequency of the variables must be the same, so the daily data are transformed into annual form. For this purpose, the annual average of each daily series is calculated, and their similarity to the Energy Security Index is measured. The objective is to select the daily time series whose historical patterns most closely mirror those of the annual index, thereby providing a robust basis for high-resolution forecasting.

Table 1 presents the five variables exhibiting the highest similarity to the Energy Security Index according to each of the six similarity methods examined in this study. As demonstrated in Table 1, the Volume_Brent time series consistently emerged as the most representative proxy: it ranked among the top five most similar variables across all methods and was identified as the most similar variable in five out of six methods. Based on this strong and consistent performance, Volume_Brent is selected as the proxy variable to stand in for the Energy Security Index in the modeling process.

Subsequently, forecasting models are developed and evaluated to predict the next 15 days of the selected proxy index. These predictive models form the foundation for generating high-frequency estimates of energy security, enabling more timely monitoring and analysis of trends and potential risks.

Table 1. Top-Ranked Proxy Variables for Energy Security Identified by Multiple Similarity Methods

| Soft-DTW Distance | DTW Distance | LCSS | edr | hausdorff |
|---|---|---|---|---|
| **Volume_Brent** | Volume_Brent | Volume_Brent | Volume_Brent | Volume_RBOB_Gasoline |
| **Volume_WTI** | Volume_WTI | Open_WTI_Oil_ETF | Volume_Heating_Oil | Volume_Brent |
| **Volume_WTI_Oil_ETF** | Volume_WTI_Oil_ETF | High_WTI_Oil_ETF | Volume_RBOB_Gasoline | Open_Oil_Services_ETF |
| **Volume_RBOB_Gasoline** | Volume_RBOB_Gasoline | Close_WTI_Oil_ETF | Open_Heating_Oil | High_Oil_Services_ETF |
| **Volume_Energy_Sector_ETF** | Volume_Energy_Sector_ETF | Adj_Close_WTI_Oil_ETF | Low_Heating_Oil | Close_Oil_Services_ETF |

**Prediction model**

XGBoost as a robust method for time series forecasting methods and as an optimized gradient boosting algorithm, is well-regarded for its ability to handle complex nonlinear relationships and deliver high predictive accuracy in time series applications. The data set is split into train and test dataset to evaluate the performance of the model. The model was trained exclusively on the training set to prevent data leakage and then evaluated on the holdout test set to assess its generalization capability. Following model training, performance metrics including the root mean squared error (RMSE), mean absolute error (MAE), and the coefficient of determination ($R^2$) were computed. The results are summarized in Table 2, providing a comprehensive overview of the model's predictive accuracy on both the training and test datasets.

Table2. the model conduction result

| | RMSE | MAE | $R^2$ |
|---|---|---|---|
| train | - | - | 0.981 |
| test | 1354.89 | 800.14 | 0.945 |

As represented in table2, training $R^2$ indicates that the model fits the training data well, capturing the underlying patterns effectively. The test $R^2$ confirms strong generalization performance, demonstrating that the model maintains predictive accuracy on unseen data. The Error value includes MAE and RMSE get 800.14 and 1354.89 which seem big value but since the target variable in this study is the Volume of Brent, it is important to consider the scale of this variable when interpreting error metrics. According to the scale of the target value, the errors value in table 2 are acceptable which confirm the adequacy of the model's predictive capabilities.

Figure 4 presents a visual comparison between the actual and predicted values on the test set. The figure shows the daily fluctuation of Brent volume as represented by blue line and orange line which consider as predicted value are almost accurate to capture daily fluctuations.

Figure4. Actual vs predicted volume_Brent

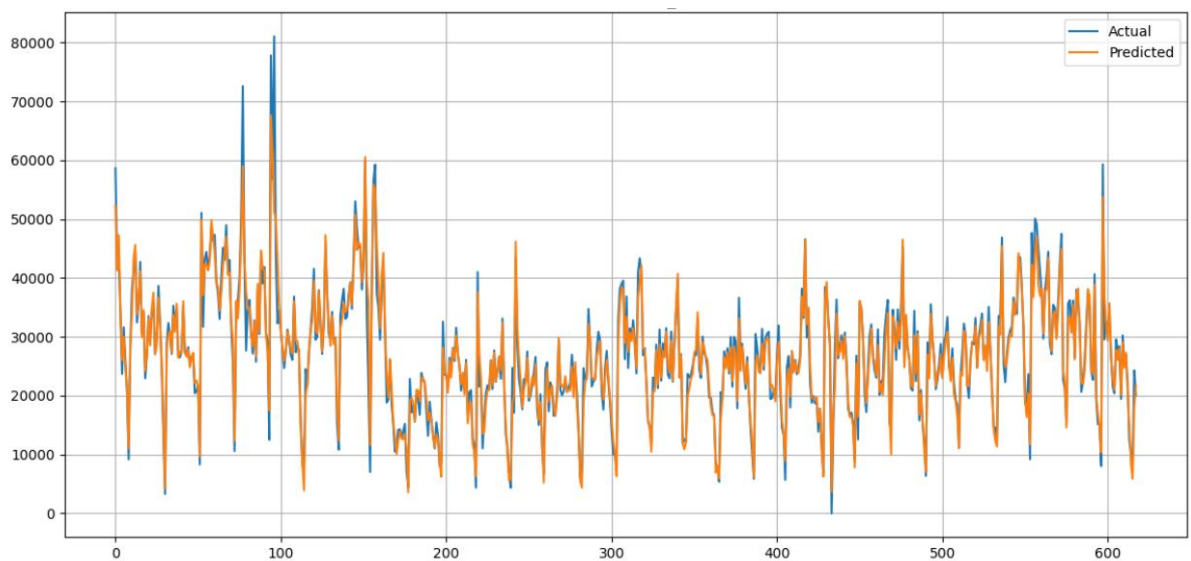

Figure 5 shows the 15 days ahead forecast of Brent_volume according to conducted model. Examining these 15 days shows the fluctuation during these days. According to the figure 2 after day 2 the target value Peaks around Day 4 and then until day 8 it Declines again. Around day 10 it rose again and until day 15 the value declined. Besides the exact value of prediction, the confident intervals are presented in figure 5.

Figure5. next 15 days prediction with 95% confidence intervals

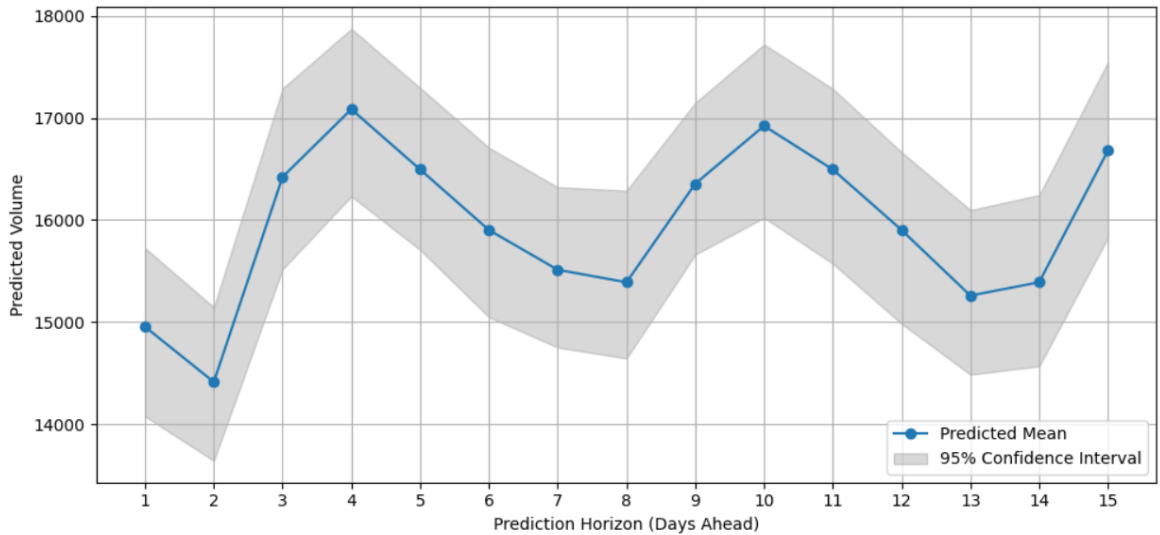

The confidence intervals represent the ranges within which the future values are expected to fall with a specified level of certainty. These intervals help quantify the uncertainty inherent in the forecasts and provide a more informative basis for interpretation and decision-making. Table 2 presents the prediction results for the next 15 days along with the corresponding confidence intervals, showing both the estimated values and the range within which each value is likely to occur.

Table 2. prediction value and intervals

|    | Predicted_Volume_Brent | Adjusted_CI_Lower_95% | Adjusted_CI_Upper_95% |
|----|------------------------|------------------------|------------------------|
| 1  | 14954.73 | 14077.26 | 15726.27 |
| 2  | 14413.91 | 13641.47 | 15145.91 |
| 3  | 16421.88 | 15518.04 | 17290.43 |
| 4  | 17085    | 16231.83 | 17873.58 |
| 5  | 16494.95 | 15706.5  | 17294.34 |
| 6  | 15902.16 | 15049.07 | 16709.2  |
| 7  | 15512.91 | 14751.36 | 16322.13 |
| 8  | 15390.29 | 14644.71 | 16288.89 |
| 9  | 16355.31 | 15661.82 | 17148.81 |
| 10 | 16921.71 | 16022.93 | 17720.95 |
| 11 | 16494.95 | 15572    | 17286.05 |
| 12 | 15902.16 | 14983.58 | 16664.58 |
| 13 | 15259.88 | 14486.25 | 16098.4  |
| 14 | 15390.29 | 14568.1  | 16245.92 |
| 15 | 16679.66 | 15826.92 | 17548.21 |

**Discussion**

The goal of this study is to predict daily energy security. Since this variable is reported on an annual basis, an essential first step was to identify a suitable daily proxy. As shown in Table 1, the Volume_Brent variable emerged as the most appropriate and robust proxy for the Energy Security Index, ranking highest in five out of the six similarity methods applied.

The next step toward producing accurate daily estimates of energy security was to build a robust predictive model. XGBoost was selected for this purpose due to its ability to model complex nonlinear relationships among predictors and its strong performance in time series forecasting tasks. As presented in Table 2, the model achieved a high training $R^2$ of 0.981, indicating an excellent fit to the training data. However, this metric alone does not fully demonstrate predictive accuracy. The $R^2$ on the test dataset confirms that the model does not suffer from overfitting and generalizes well to unseen data.

Considering the scale of the target variable (Volume_Brent), the resulting RMSE (1,354.89) and MAE (800.14) are both within acceptable ranges, supporting the conclusion that the model performs effectively.

The model was then used to generate forecasts 15 days ahead for Volume_Brent as a proxy for the Energy Security Index. The predicted values fluctuate between approximately 14,000 and 17,000 over the forecast horizon. As illustrated in Figure 2, the proxy variable is expected to exhibit periodic short-term volatility during this period.

Prediction uncertainty is expressed through confidence intervals, which indicate the range within which the actual proxy values are expected to fall with 95% probability. The intervals widen somewhat around peaks and troughs, reflecting increased uncertainty during more volatile periods. Overall, the width of the intervals remains moderate, suggesting a reasonable degree of prediction certainty.

Importantly, the intervals have been broadened to account for the known discrepancy between the proxy variable and the actual Energy Security Index. This means the intervals reflect not only model uncertainty (the error inherent in predicting Volume_Brent itself) but also the additional uncertainty arising from using a proxy to approximate daily energy security. This approach provides a more conservative and realistic estimate of prediction uncertainty.

Because this variable serves as a proxy for the Energy Security Index, the forecasts can be interpreted as indicative of the expected near-term behavior of energy security, assuming that the historical similarity relationship between the two variables remains stable. The results suggest a period of moderate fluctuations, with no abrupt spikes or collapses anticipated over the next two weeks.

**Conclusion**

This study proposes a novel methodology for bridging the temporal gap between low-frequency and high-frequency data by identifying an appropriate daily proxy for annually reported indicators—in this case, the Energy Security Index. Through the application of advanced time series similarity methods, including DTW, Soft-DTW, LCSS, EDR, and Hausdorff distance, the study identifies *Volume_Brent* as the most reliable proxy variable. This identification is based on its consistent top ranking across multiple similarity metrics, reflecting its close temporal alignment with the target index. The predictive phase of the study leverages the XGBoost algorithm, which is particularly well-suited for capturing complex nonlinear patterns in time series data. Model evaluation metrics such as $R^2$ (0.945 on test data), RMSE, and MAE demonstrate that the trained model maintains high accuracy and generalization capability. These outcomes validate the feasibility of using high-frequency proxy variables, in combination with robust machine learning models, to estimate and forecast otherwise infrequently reported macroeconomic indicators. By forecasting 15-day ahead values of *Volume_Brent*, the study not only presents a granular view of the proxy's expected behavior but also introduces confidence intervals that incorporate both model uncertainty and proxy-based approximation error.

This dual-layered approach to uncertainty quantification offers a more realistic and cautious perspective on forecast interpretation. Ultimately, this research contributes a scalable and practical framework for real-time monitoring of critical economic and energy indicators using high-frequency data. It enables policymakers and analysts to respond to market signals and policy outcomes in a timelier manner, which is essential in fast-changing and volatile environments such as the global energy market. Future studies could further refine this approach by integrating multiple proxies, employing ensemble models, or extending the method to other policy-relevant macroeconomic variables.